\documentclass[sigconf]{acmart}

\AtBeginDocument{%
  \providecommand\BibTeX{{%
    \normalfont B\kern-0.5em{\scshape i\kern-0.25em b}\kern-0.8em\TeX}}}

\setcopyright{acmcopyright}
\copyrightyear{2024}
\acmYear{2024}
\acmDOI{XXXXXXX.XXXXXXX}

\usepackage{xcolor}
\usepackage{url}
\usepackage{comment}
\usepackage{graphicx}
\usepackage{caption}
\usepackage{subcaption}
\usepackage{multirow}
\usepackage{svg}
\usepackage{adjustbox}
\usepackage{xcolor,colortbl}
    \definecolor{Gray}{gray}{0.85}
\usepackage{cleveref}
\usepackage{xcolor,soul}

\renewenvironment{quote}%
  {\list{}{\leftmargin=0.2in\rightmargin=0.2in}\item[]}%
  {\endlist}

\acmConference[HRI '24 (LBR Accepted Manuscript)]{The ACM/IEEE International Conference on Human-Robot Interaction}{March 11--14, 2024}{Boulder, CO, USA}
\acmPrice{15.00}
\acmISBN{978-1-4503-XXXX-X/18/06}

\setcopyright{none}
\renewcommand\footnotetextcopyrightpermission[1]{} 
\begin{document}

\title{The Effect of Predictive Formal Modelling at Runtime on Performance in Human-Swarm Interaction}


\author{\large Ayodeji O. Abioye$^1$, William Hunt$^1$, Yue Gu$^2$, Eike Schneiders$^3$, Mohammad Naiseh$^4$, Joel E. Fischer$^3$, \\Sarvapali D. Ramchurn$^1$, Mohammad D. Soorati$^1$, Blair Archibald$^2$, and Michele Sevegnani$^2$}

\thanks{$^1$Electronics and Computer Science, University of Southampton, UK}
\thanks{$^2$School of Computing Science, University of Glasgow, UK}
\thanks{$^3$School of Computer Science, University of Nottingham, UK}
\thanks{$^4$Computing and Informatics, Bournemouth University, UK}

\renewcommand{\shortauthors}{Abioye et al.}

\begin{abstract}
Formal Modelling is often used as part of the design and testing process of software development to ensure that components operate within suitable bounds even in unexpected circumstances. In this paper, we use predictive formal modelling (PFM) at runtime in a human-swarm mission and show that this integration can be used to improve the performance of human-swarm teams. We recruited 60 participants to operate a simulated aerial swarm to deliver parcels to target locations. In the PFM condition, operators were informed of the estimated completion times given the number of drones deployed, whereas in the No-PFM condition, operators did not have this information. The operators could control the mission by adding or removing drones from the mission and thereby, increasing or decreasing the overall mission cost. The evaluation of human-swarm performance relied on four key metrics: the time taken to complete tasks, the number of agents involved, the total number of tasks accomplished, and the overall cost associated with the human-swarm task.  Our results show that PFM modelling at runtime improves mission performance without significantly affecting the operator's workload or the system's usability.
\end{abstract}

\ccsdesc[500]{Human-centered computing~Empirical studies in interaction design}
\ccsdesc[500]{Human-centered computing~User interface design}
\ccsdesc[500]{Human-centered computing~User studies}
\ccsdesc[100]{Human-centered computing~Collaborative interaction}
\ccsdesc[100]{Human-centered computing~Graphical user interfaces}

\keywords{Human-Robot Interaction (HRI), Human-Swarm Interaction (HSI), Predictive Formal Modelling (PFM), Task Performance}

\settopmatter{printfolios=true}

\maketitle

\section{Introduction}
Aerial swarms amplify our ability to observe and engage with areas that are challenging for us to reach or oversee. One of the promising applications of aerial swarms is in search and rescue missions to locate and identify casualties on time or to deliver essential life-saving supplies to remote and difficult-to-reach areas in the aftermath of a natural disaster~\cite{abioye2023}. Such applications are often accompanied by several challenges, ranging from design and deployment~\cite{Hoang2023} to issues related to safety~\cite{Lee2022}, regulations~\cite{Lyu2023}, and the operator's mental workload~\cite{Abioye2022}. Other challenges include performance, shared control and degree of automation, 
as well as determining the appropriate timing for presenting necessary information to the human operator. Prior research on human-swarm interaction (HSI) has identified essential prerequisites for the successful operation of aerial swarms~\cite{Clark2022}. For systems relying on human supervision and intervention, a critical requirement for the smooth operation of the swarm is the efficient timing and selection of relevant data provided to the operator.~\citet{Gu2023} proposed a predictive formal modelling (PFM) technique to estimate the mission success at runtime. PFM can be used to inform the swarm operator about the crucial information required to make appropriate decisions in order for the mission to be successful. This capability allows the swarm operator to make informed decisions promptly, increasing the overall efficiency and adaptability of the human-swarm collaborative efforts. With PFM providing crucial insights, the interaction becomes more streamlined and responsive, ensuring that decisions are aligned with the mission's success criteria. This enhanced decision support contributes to a more seamless and effective collaboration between humans and robots in achieving mission objectives. Our hypothesis posits that incorporating predictive formal modelling (PFM) into the real-time execution of human-swarm tasks can empower swarm operators to make more informed decisions, thereby optimising the utilisation of swarm capabilities and resources.

In this paper, we integrate PFM into the `Human And Robot Interactive Swarm' (HARIS) simulator~\cite{hunt2023demonstrating} to provide human swarm operators with real-time mission and swarm status updates, along with predictions of mission success. 
Following a within-subject design, we recruited 60 participants to complete a human-swarm task of delivering packages to different areas with two conditions (with and without PFM).  We assess the impact of PFM on the performance and required mental workload during the completion of a task as a human-swarm team. Results of the study showed that the PFM condition was able to significantly improve mission performance without having a significant effect on the operator's workload.

\section{Related Work}

In~\citet{abioye2023}, the authors evaluated the effect of adding an extra feature to the human-swarm interface, i.e. operator option to request high-quality images of a search area. They found that this led to higher trust perception but did not enhance the overall human-swarm performance. \citet{Schneiders:2022:THRI} indicated the demand for studying non-dyadic human-in-the-loop system configurations, such as that presented in this work. \citet{hunt2023demonstrating} proposed a method of dynamic re-tasking and triage based on operator feedback. 
~\citet{Wilson:2023:Swarm} identified key challenges for the deployment and use of robot swarms, which included how humans understand, monitor, control, and interact with swarms. 

\citet{kouvaros2015verifying} show a formal model for swarms and determine whether the emergent behaviour is satisfied; \citet{boureanu2016verifying} analyse unbounded swarm systems with respect to security requirements by verifying a parameterised model; \citet{lomuscio2021verifying} outline a verification procedure to reason about the fault-tolerance in probabilistic swarm systems. However, none of these approaches can give guarantees after deployment. A close approach to ours is runtime monitoring~\cite{bartocci2018introduction}, where pre-constructed monitors are used to analyse the system execution traces that are generated at runtime against formal specifications~\cite{falcone2021taxonomy}. These monitors can evolve with system dynamics, such as the size and topology, but cannot reason about mission-level specifications, like the human-swarm interactions, where finite observations are not sufficient. Instead, \citet{Gu2023} propose a framework to integrate runtime modelling~\cite{blair2009models}, that has been deployed in system reasoning for unforeseen situations during execution~\cite{bencomo2012view}, with formal methods and focus on formal runtime modelling. Quantitative formal models can provide predictions, and this has been used at design time, e.g., for predicting failures and service availability of components~\cite{DBLP:journals/tdsc/CalderS19}. In this work, we adopt an existing model~\cite{Gu2023} implementing PFM at runtime, to predict the feasibility of a human-swarm mission succeeding and check whether presenting PFM output in the user interface can support human operator in their decision-making during the human-swarm task.

\section{Study}
We conducted a within-subject user study with 60 participants divided into two counterbalanced groups as shown in Table~\ref{tab:Study}, in order to directly compare usability, workload, and performance between the PFM and No-PFM conditions.

\begin{table}[!htb]
\small
\renewcommand{\arraystretch}{1.3}
\caption{Showing counterbalanced distribution of the 60 recruited participants.}
\begin{tabular}{|c|l|l|l|}
\hline
\rowcolor[HTML]{EFEFEF} 
Group & Scenario 1 & Scenario 2 & \# \\
\hline
1 & No Prediction (No-PFM) & With Prediction (PFM) & 30 \\
\hline
2 & With Prediction (PFM) & No Prediction (No-PFM) & 30 \\ 
\hline
\end{tabular}
\label{tab:Study}
\end{table}

\subsection{Participants}
We recruited 60 participants ($38$ female, $22$ male, average age: $34.6$, age range $18$ - $64$) through Prolific~\cite{prolific}\footnote{We received ethics approval from the University of Southampton's ethics committee (confirmation number: ERGO/FEPS/85523).}. $70 \%$ of participants have at least a bachelor's degree, $50 \%$ were above-average computer users, and $47 \%$ were familiar with UAV or swarm robotics. Participants were recruited from the US and the UK. Participants were randomly allocated into groups and received £ $9$. The average study duration was $33.6$ minutes.

\subsection{Study task}
\begin{figure}[tb]
        \centering
        \framebox{\includegraphics[width=0.93\columnwidth]{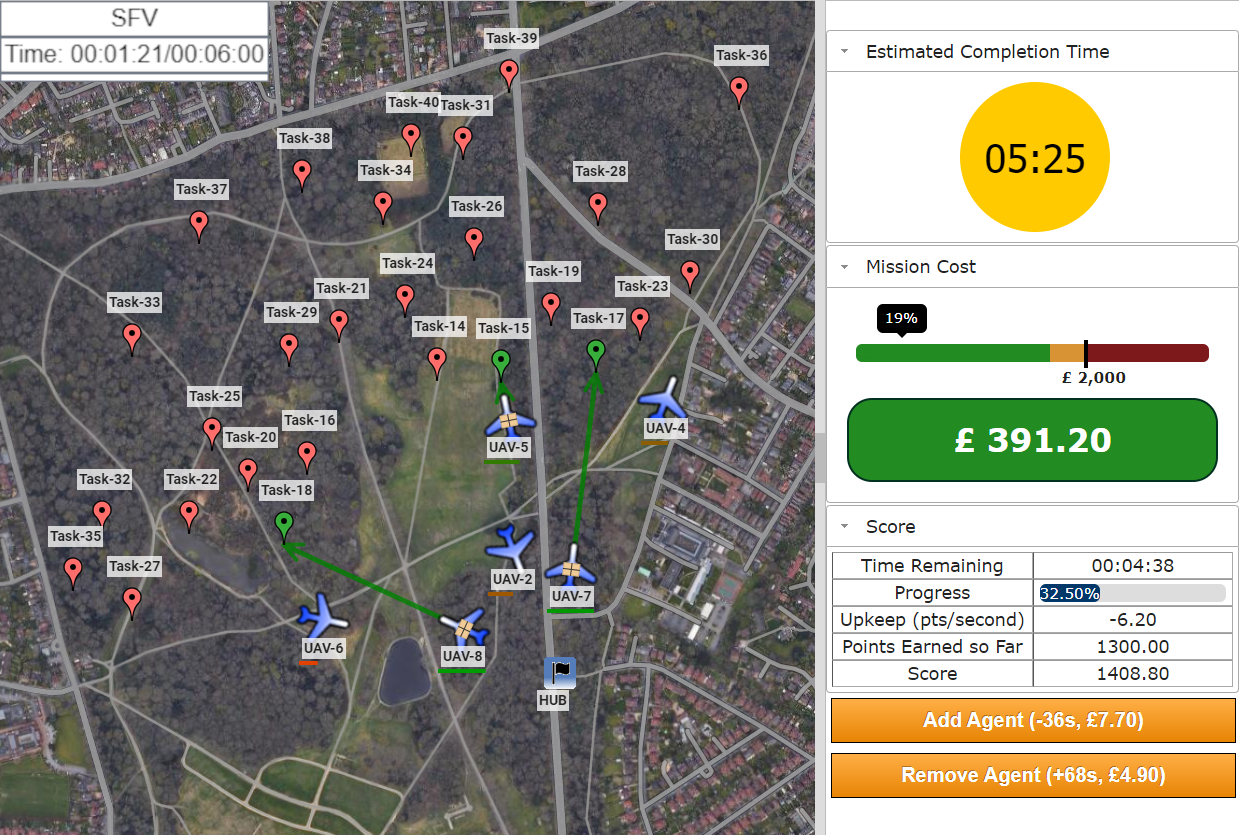}}
    \caption{HARIS simulator interface showing the predictive formal modelling based estimated completion time feature in the PFM scenario.}
    \label{fig:haris_with_pfm_time_estimation}
\end{figure}

To investigate the impact of PFM, in a human-swarm interaction context, we developed a drone delivery mission scenario which was presented to participants through two scenarios: the PFM and No-PFM scenarios. 
To constrain the operators' strategies, we added a cost (£ 2,000) and time (6 minutes) limit. These two constraints meant that the participants were occupied trying to meet the time limit while staying within the mission budget. Users could control the swarm and the mission by performing two operations: adding or removing UAVs from the mission. The more UAVs they add, the faster they complete the mission, but they incur a higher mission cost. The reverse is also applicable in that the more UAVs they remove from the mission, the longer the mission completion time, and the less the overall mission cost. We set the minimum and maximum number of UAVs allowed in the mission to be 4 and 10 respectively. The final mission cost was a cumulative sum of the upkeep cost per second. We implemented a non-linear per-second upkeep cost function that makes the upkeep per-second cost higher each time a new UAV is added. This reduces the participant's ability to predict the cost of adding or removing a UAV, especially for the No-PFM condition without the predictive model. We added 40 delivery tasks to each scenario. PFM was used to predict the probability of completing all deliveries with the given number of UAVs. This prediction was presented to the operator as an estimated completion time, as shown inside the circle in the upper right corner (see~\Cref{fig:haris_with_pfm_time_estimation}). The colour of the circle changes from green to yellow and red depending on the estimated time of completion (green for finishing well below the given time, yellow for finishing near the given time, and red for exceeding it).

\subsection{HARIS Model}
The HARIS simulator is a browser-based platform that was specifically designed for human-in-the-loop multi-agent and swarm robotics experimentation. HARIS is a successor of HutSim~\cite{ramchurn2015study} which was designed with a specific focus on usability by consulting with industry experts to model not only their typical command structure but also make it operable with real-life or simulated agents. Building on its predecessor, HARIS was further tailored to its use case derived from interviews with drone pilots~\cite{parnell2022trustworthy} and swarm experts~\cite{naiseh2023outlining} to make the platform as usable and realistic as possible while maximising the ease of use for multiple human operators~\cite{soorati2022enabling}, making this simulator a useful tool for the investigations on human-swarm simulations.

We use the model from~\cite{Gu2023} with slight modifications to better reflect the scenario. For example, the background failure in each region is removed to relieve participants’ stress from the unexpected loss of UAVs; Erlang-k law~\cite{erlang1917solution} is implemented to represent a smooth transition delay in CTMCs. The integration with the HARIS simulator follows a similar process to~\cite{Gu2023}, but with the Sim2PRISM middleware directly embedded in HARIS. Instead of showing the probability of success directly, which might be difficult for participants to interpret, we consider the feasibility over different time intervals and give an estimated completion time as the time when the probability of success reaches $0.99$. Additionally, we implement What-if scenarios to give the participants extra information on the effect of adding/removing a UAV before making a decision.

\subsection{Procedure}\label{subsec:procedure}
Following recruitment, participants were presented with the participant information sheet and consent form, after which they completed a brief demographics survey which collected data on gender, age group, education level, self-rated computer expertise, as well as self-rated UAV or Swarm robotics knowledge. Subsequently, participants were asked to watch a short study briefing video and asked to answer three questions to test their preliminary understanding of the task. To ensure that participants understood the study task, two of these three had to be answered correctly in order to proceed with the study. Participants were required to perform a short tutorial scenario which allowed them to experiment with all the provided functionality and experience the interface prior to the actual data collection. Participants then proceeded to their first scenario. Following its completion, they completed the post-task survey which included the 6-item NASA-TLX~\cite{hart1986nasa} questionnaire and the 10-item System Usability Scale (SUS)~\cite{Brooke1995SUS}. They continued with the second scenario, followed by the same set of questionnaires. Finally, participants were asked to complete a short survey in relation to their preferred scenario condition, before returning to Prolific. These questions were related to (a) the perceived accuracy of the time estimation feature provided (for the PFM condition), (b) their preferred scenario, (c) a selection of reasons for perceived success during task completion, (d) the primary reason for their success, and (e) a binary selection if they used the estimated completion time. Each participant's performance was measured and recorded in real-time during the scenario tasks as HARIS generated log files.

The survey questionnaires and HARIS simulator were dockerised and deployed online\footnote{Online HARIS simulator: \url{https://uos-haris.online/}} on an AWS EC2 c5.4xlarge (32GB RAM, 16 vCPUs) instance running the Ubuntu 22.04 operating system. The dockerisation was necessary for a scalable deployment due to the high computing resource requirement of the prediction model in the simulator.

\section{Results and Analysis}
\begin{figure*}[!htb]
    \centering
    \begin{subfigure}[b]{0.32\textwidth}
        \centering
        \adjustbox{trim=1cm 0cm 1.2cm 0cm}{
          \includegraphics[width=0.99\textwidth]{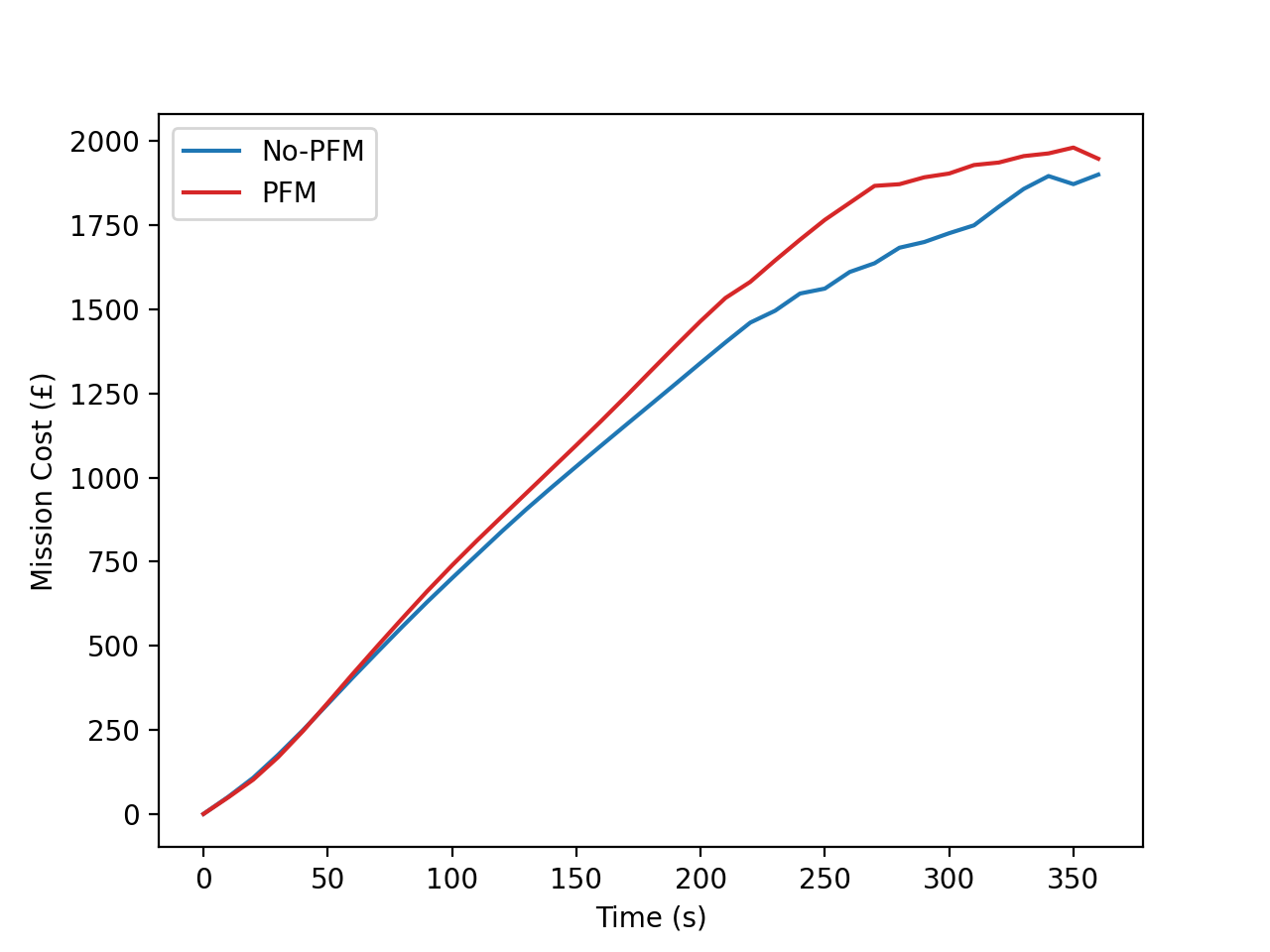}
        }
        \caption{Mission Cost}
        \label{fig:mission_cost_all}
    \end{subfigure}
    \hfill
    \begin{subfigure}[b]{0.32\textwidth}
        \centering
        \adjustbox{trim=1cm 0cm 1.2cm 0cm}{
            \includegraphics[width=0.99\textwidth]{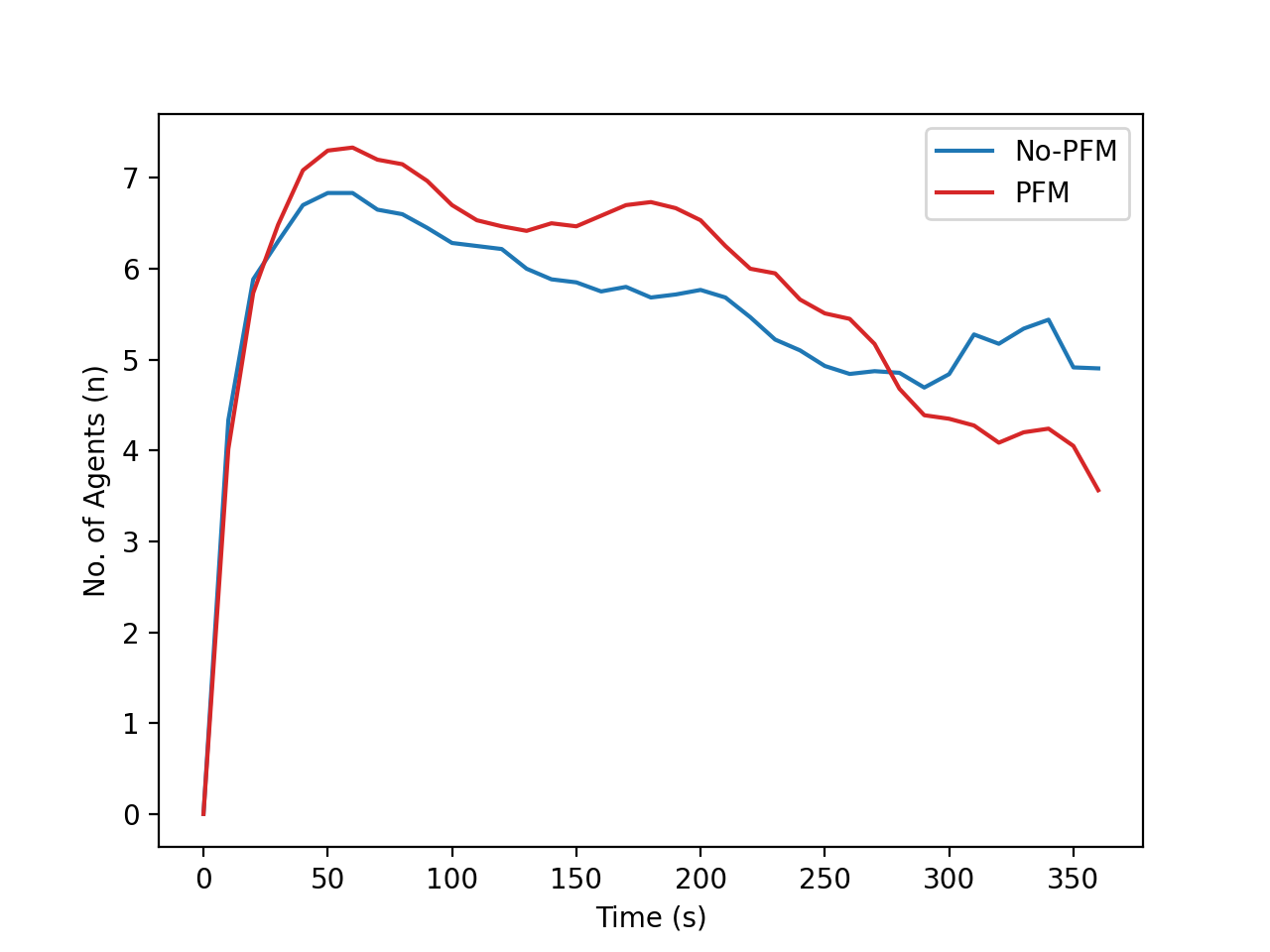}
        }
        \caption{Number of Agents}
        \label{fig:num_agents_all}
    \end{subfigure}
    \hfill
    \begin{subfigure}[b]{0.32\textwidth}
        \centering
        \adjustbox{trim=1cm 0cm 1.2cm 0cm}{
            \includegraphics[width=0.99\textwidth]{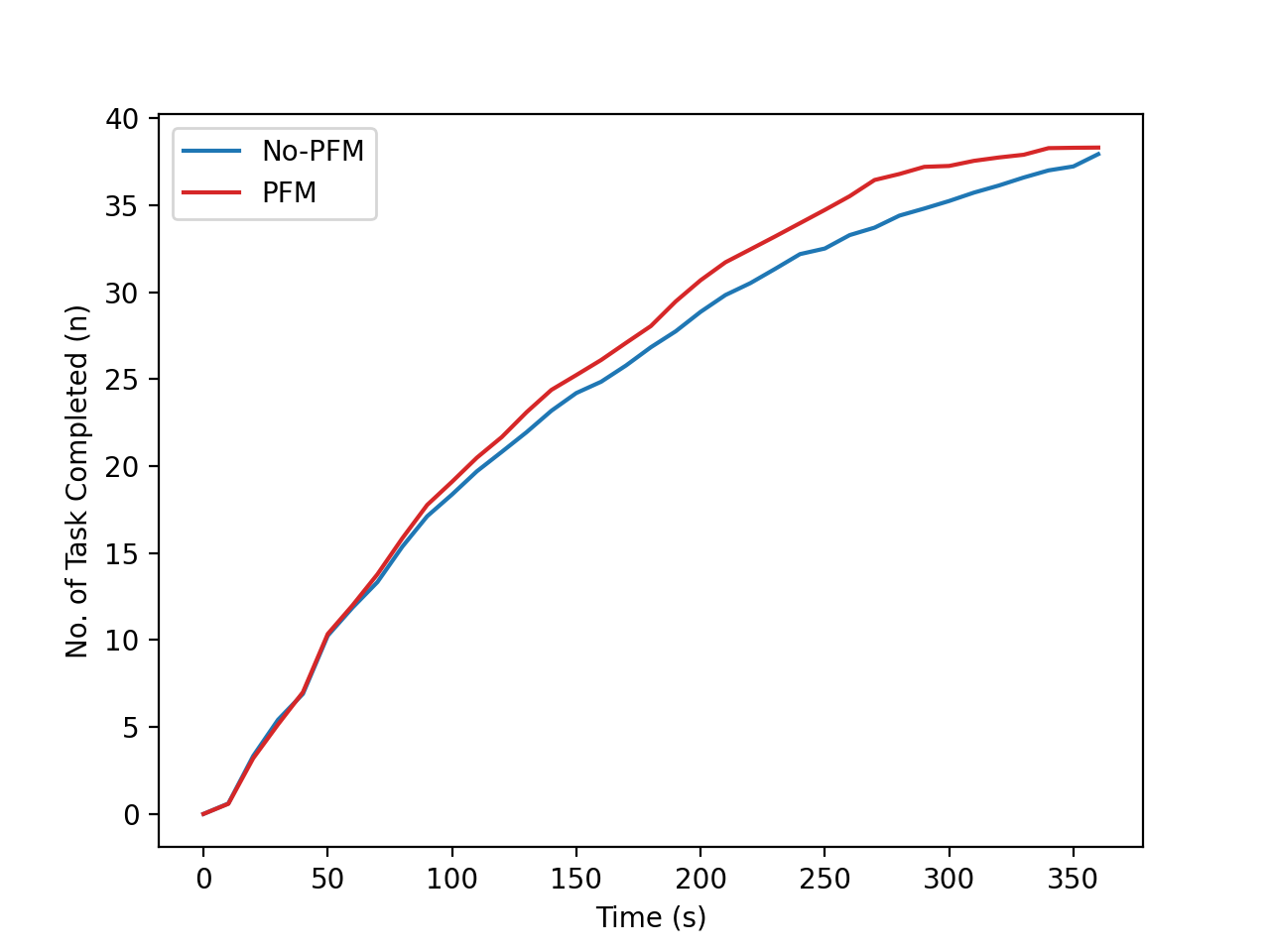}
        }
        \caption{Completed Tasks}
        \label{fig:completed_task_all}
    \end{subfigure}
    \caption{Comparing the mean performance of all study conditions over time.}
    \label{fig:mean_performance_all}
\end{figure*}

The result of the participants' performance over time is presented in Figure \ref{fig:mean_performance_all}. Figure \ref{fig:mission_cost_all} shows the mean mission cost of each scenario over time. The No-PFM scenario incurred a lower cost over time than the PFM scenario. Figure \ref{fig:num_agents_all} shows the mean number of agents used over time. The No-PFM group started with the least number of agents but finished with the most. Since this group did not have the estimated completion time displayed, it is possible that they realised very late that they may not finish, and therefore started adding more agents towards the end. This might indicate that participants in the No-PFM condition found it more difficult to balance the number of agents with the two constraints defined. Figure \ref{fig:completed_task_all} compares the mean number of completed tasks over time and shows that participants completed more tasks on average over time in their PFM scenario compared to their No-PFM scenario.

To determine the impact of our formal model on the performance of the human-swarm teaming, we analysed the results to understand whether the prediction feature contributes to participants completing tasks more efficiently, with minimal influence on the overall mission cost. We evaluated four dependent variables: a) \texttt{Time Completion}: refers to the mission completion time i.e. the time taken to complete 40 delivery tasks. b) \texttt{No. of Agents}: refers to the mean number of agents deployed by each participant to complete the delivery task. c) \texttt{Completed Tasks}: We considered a delivery task to be successfully completed when the UAV reaches the target coordinate. After this, the UAV returns to the hub to collect parcels for the next delivery. d) \texttt{Cost per Task}: This was computed as a ratio of the mean total cost incurred over the mean number of tasks completed per study scenario. 

\begin{table}[h]
\small
\centering
\caption{Descriptive statistics and one-way ANOVA results. Significance levels are indicated as: * $p < 0.05$ , ** $p < 0.01$}
\label{table:experiment1_performance_result}
    \begin{tabular}{|c|c|l|c|c|l|}
        \hline
        \rowcolor[HTML]{EFEFEF} 
        Variable & Scenario & Mean& Std. & F value & p value\\
        \hline
        Time Completion \multirow{2}{*}& No-PFM&329s&36.93&5.363&0.022*\\
                            &PFM&314s&34.77&&\\
        \hline
        No. of Agents \multirow{2}{*}& No-PFM&5.79&1.06&3.074&0.082\\
                            &PFM&6.10&0.86&&\\
        \hline
        Completed Tasks\multirow{2}{*}& No-PFM&38.80&1.71&7.255&0.008**\\
                            &PFM&39.55&1.30&&\\
        \hline
        Cost per Task\multirow{2}{*}& No-PFM&£52.85&7.31&0.001&0.988\\
                            &PFM&£52.83&3.80&&\\
        \hline
    \end{tabular}
\end{table}

In terms of data analysis, our dataset meets the prerequisites necessary for conducting one-way ANOVA testing. Additionally, we performed a G*Power analysis to verify that our sample size aligns with the required criteria. 
Specifically, we have 60 participants across experimental groups, with an assumed effect size of 0.2 and a significance level of 0.05. The G*Power analysis helps to confirm that our study is adequately powered to detect the expected effects.

For workload, participants had a mean of 4.77 (SD =1.50) in the PFM and a mean of 4.74 (SD = 1.56) in the No-PFM scenarios. One-way ANOVA for workload revealed no significant main effect (F(1, 118) = 0.009, p = 0.924. This suggests that the PFM feature did not add extra workload to the participants. Regarding usability, we used the system usability scale (SUS) to compare the mean values for the two conditions. In line with the guidelines~\cite{Brooke1995SUS,sus68}, interfaces with a value of 68 or above are considered good. Mean SUS scores for PFM and No-PFM scenarios were 70.75 (SD = 17.52) and 74.38 (SD = 15.15). This shows that the usability of both systems was good. One-way ANOVA yielded no significant effect on usability (F(1, 118) = 1.470, p = 0.228). This suggests that the PFM feature did not make the system more or less usable than without it.

As depicted in Table \ref{table:experiment1_performance_result}, the PFM condition led to enhanced task completion rates and reduced time requirements when compared to the No-PFM scenario where no prediction was presented to participants. Specifically, participants, on average, completed 39.55 tasks (SD No. of Tasks = 1.30) within 314 seconds (SD Time Completion = 34.77) in the PFM condition. This performance contrasted with the No-PFM condition, where participants completed an average of 38.80 tasks (SD No. of Tasks = 1.71) in 329 seconds (SD Time Completion = 36.93). An ANOVA test was conducted and showed that this difference was significant both in terms of Time Completion [F(1,59)= 5.363, p= 0.022*] and No. of Completed Tasks [F(1,59)= 7.255, p= 0.008*].
Moreover, our findings indicate that employing PFM prediction did not influence the utilisation of additional agents or the associated task cost in the context of human-swarm collaboration when compared to scenarios without prediction (No-PFM). In the PFM condition, participants, on average, employed 6.10 agents (SD No. of Agents = 0.86) at an average cost of £52.83 (SD Cost = 3.80) per task. Conversely, in the No-PFM condition, participants used an average of 5.79 agents (SD No. of Agents = 1.06) at a cost of £52.85 (SD Cost = 7.31) per task. An ANOVA test was conducted and showed that this difference between PFM and No-PFM was not significantly different for No. of Agents [F(1,59) = 3.074, p=0.082] and Cost per task [F(1,59)= 0.988]. Figure \ref{fig:mission_cost_all} shows the mean mission cost of each scenario over time. 

The significant results in relation to mission completion times were also reflected in the anecdotal open-ended statements made by participants following both conditions. Participants indicated the perceived usefulness of the features provided in the PFM condition as, e.g., expressed by P46:
\begin{quote}
    \textit{``I found the presence of the estimated completion time feature [PFM] helped me decide whether to add or remove agents, whereas in the first scenario [No-PFM] I was trying to estimate it myself based on the remaining time and the percentage completion of the task.''} - P46
\end{quote}
indicating the usefulness of the additional information provided to complete the task successfully. A similar sentiment was presented by P20 who describes the use of the PFM feature as a guiding mark for optimising the addition and removal of drones.
\begin{quote}
    \textit{``I used the estimated time to allow me to hover around the 6-minute mark, adding and taking away planes where necessary''} - P20
\end{quote}

\section{Discussion and Future Works}
We found that there was no significant change in workload between the two conditions and both the PFM and No-PFM interfaces were found to be usable based on the systems usability survey. Although our result show that there is a performance gain when using the predictive formal modelling feature at runtime, our analysis does not take into account how the accuracy of the prediction could affect the users' performance or trust in the system. Furthermore, this study did not investigate how the PFM feature increases explainability and hence trust in the human swarm interaction. A follow-up study could collect data on trust, acceptability, and user preferences in order to evaluate these measures. In future work, we may also consider embedding recommendations for the operators to help in controlling the swarm, i.e., when to add or remove drones. In order to expand on the assessment of mental workload, different data streams about the users' interaction, such as neurophysiological responses (e.g., error potentials), might be useful. This could indicate the operator's cognitive workload and situational awareness as they operate the swarm in a disaster response scenario.

\section{Conclusion}
Building on previous work in human-swarm interaction on deploying predictive formal modelling (PFM) at runtime, we conducted a within-subject user study to determine its impact on performance and mental workload. We recruited 60 participants to perform the role of a UAV swarm operator facilitating the delivery of parcels to target locations in a simulation environment. The role required the participant to add or remove agents as needed to complete the mission within the given time. We find that participants using PFM were able to complete more tasks in less time compared to the No-PFM scenario. This increase did not result in a higher demand on mental workload, showing the potential benefit of predictive formal modelling in a human-swarm interaction scenario.

\begin{acks}
The authors wish to acknowledge the support received from the UKRI Trustworthy Autonomous Systems Hub (EP/V00784X/1), the EPSRC project on Smart Solutions Towards Cellular-Connected Unmanned Aerial Vehicles System (EP/W004364/1), and an Amazon Research Award on Automated Reasoning.
\end{acks}

\bibliographystyle{ACM-Reference-Format}
\balance
\bibliography{ref}

\end{document}